\documentclass[11pt,a4paper]{article}
\usepackage{acl2020}
\usepackage{times}
\usepackage{latexsym}
\usepackage[inline]{enumitem}
\usepackage{multirow}
\usepackage{graphicx}
\usepackage{makecell}

\newcommand\blfootnote[1]{%
  \begingroup
  \renewcommand\thefootnote{}\footnote{#1}%
  \addtocounter{footnote}{-1}%
  \endgroup
}

\usepackage{microtype}

\aclfinalcopy 

\title{Restructuring Conversations using Discourse Relations for Zero-shot Abstractive Dialogue Summarization}

\author{Prakhar Ganesh $^*$ \\
  \texttt{prakhar.g@adsc-create.edu.sg} \\\And
  Saket Dingliwal $^*$ \\
  \texttt{sdingliw@cs.cmu.edu} \\}

\date{}

\begin{document}
\maketitle

\begin{abstract}
Dialogue summarization is a challenging problem due to the informal and unstructured nature of conversational data. Recent advances in abstractive summarization have been focused on data-hungry neural models and adapting these models to a new domain requires the availability of domain-specific manually annotated corpus created by linguistic experts. We propose a zero-shot abstractive dialogue summarization method that uses discourse relations to provide structure to conversations, and then uses an out-of-the-box document summarization model to create final summaries. Experiments on the AMI and ICSI meeting corpus, with document summarization models like PGN and BART, shows that our method improves the ROGUE score by up to 3 points, and even performs competitively against other state-of-the-art methods.
\blfootnote{$^*$ equal contribution}
\end{abstract}

\section{Introduction}
\label{sec:introduction}

With the increase in generation of different forms of textual information, auto summarization tools have gained popularity. A significant portion of this information occurs in the form of conversations between multiple participants, for example, email threads, social network comments, meeting transcripts, \textit{etc.}, which emphasizes the need for a dialogue summarization framework. Two popular forms of summarization are extractive and abstractive. The former assembles the summary from the source text directly by choosing relevant sentences while the latter generates novel words and sentences in the final summary. Abstractive summaries are more intuitive to read and resembles the human approach to summarization.

Recent progress in abstractive summarization for structured documents has gained significant attention \cite{see2017get, tan2017abstractive, hoang2019efficient, lewis2019bart}. This can be attributed to the availability of large datasets, like the CNN/Dailymail dataset \cite{hermann2015teaching, nallapati2016abstractive}, which facilitates the training of complex neural models for the task.

However, abstractive dialogue summarization is a challenging task mainly because \begin{enumerate*}[label=(\roman*)]
  \item Dialogues generally have multiple speakers and inherent relations across utterances, for example, question-reply pairs, that needs to be appropriately modeled.
  \item Dialogues can contain informal phrases and pause fillers like \textit{"umm"}, \textit{"uhh"}, \textit{etc.} which makes it difficult to process it in the same way as document summarization.
  \item There are no large publicly available annotated datasets or benchmarks for abstractive dialogue summarization that can be used to train data-hungry deep generative models.
\end{enumerate*}

Inspired by these challenges, we propose a novel two-phase pipeline for dialogue summarization which does not require any training data.
We exploit discourse relations \cite{stone2013situated, qin2017joint} to restructure conversations into a document in the first phase, followed by a document summarization model \cite{see2017get, lewis2019bart} to generate the final summary in the second phase.
The major benefits of using this two-phase approach includes the flexibility of using any document summarization model,
no requirement of end-to-end annotated training data, interpretability, reduced complexity and ease of debugging.

The main contribution of the paper are as follows:
\begin{enumerate*}[label=(\roman*)]
  \item Unlike most studies, we solve the problem of abstractive dialogue summarization from the perspective of zero-shot learning, which is often the case in this particular application.
  \item We propose a novel two-phased pipeline which combines existing models and exploits the performance of the well-explored field of document summarization.
  \item We provide extensive experimentation to show that our method consistently improves the performance of out-of-the-box document summarizers, and can even beat state-of-the-art results with a good base document summarization model.
\end{enumerate*}

\section{Related Work}
\label{sec:related_work}
Automatic document summarization has been well studied in the past and with the advent of deep generative language models, abstractive summarization is gaining momentum. 
Various models that do abstractive summarization includes RNN models \cite{rush2015neural, chopra2016abstractive, see2017get}, graph-based models \cite{tan2017abstractive}, pre-trained transformer models \cite{hoang2019efficient, lewis2019bart} and reinforcement learning (RL) models \cite{chen2018fast}.


Prior works for dialogue summarization mostly followed extractive approaches, for example, \citet{garg2009clusterrank} used graph-based methods (ClusterRank) while \citet{galley2006skip} used skip-chain CRFs to rank utterances based on importance. 
However, with recent success of seq2seq models, the focus of research has shifted to abstractive summarization in dialogues \cite{goo2018abstractive, yuan2019abstractive, liu2019topic, pan2018dial2desc, zhao2019abstractive, shang2018unsupervised}. However due to the absence of huge amounts of training data, most of these methods resort to learning topic descriptions instead of summaries, to train and evaluate their models \cite{goo2018abstractive, pan2018dial2desc}. These topic descriptions are far more concise and general, and does not capture speaker specific information and the flow of the conversation. 

Methods like \citet{yuan2019abstractive, liu2019topic} use additional human-annotated information like dialog domain, topic segments etc., specific to their own dataset, which might not be available in real-life conversations. In contrast, here we predict discourse labels which can be generalised for any form of conversation.
A closely related work by \citet{goo2018abstractive} also leverages discourse labels as features in a neural summarization model. However, we instead use discourse labels to restructure the dialogue. Additionally, as mentioned earlier, they train and evaluate their model on topic descriptions and not on longer summaries.

\citet{shang2018unsupervised} uses a complex four-phased unsupervised approach for abstractive summarization using pre-trained word vectors and language model. These phases introduce a number of parameters which needs to be tuned on domain-specific data. Thus, while they do not need any labelled data for training, they are still one step away from generalizing to out-of-domain conversations. Further, they rely on certain redundancies in the spoken dialogues which might not be general to every conversation. 
However, we use discourse relations to restructure the dialogue into a document and then leverage the power of state-of-the-art document summarization models to significantly simplify the pipeline, and do not require any form of domain-specific data, labelled or unlabelled.

\section{Methodology}
\label{sec:proposed_solution}

\begin{figure*}[]
\centering
\includegraphics[width = 1\textwidth]{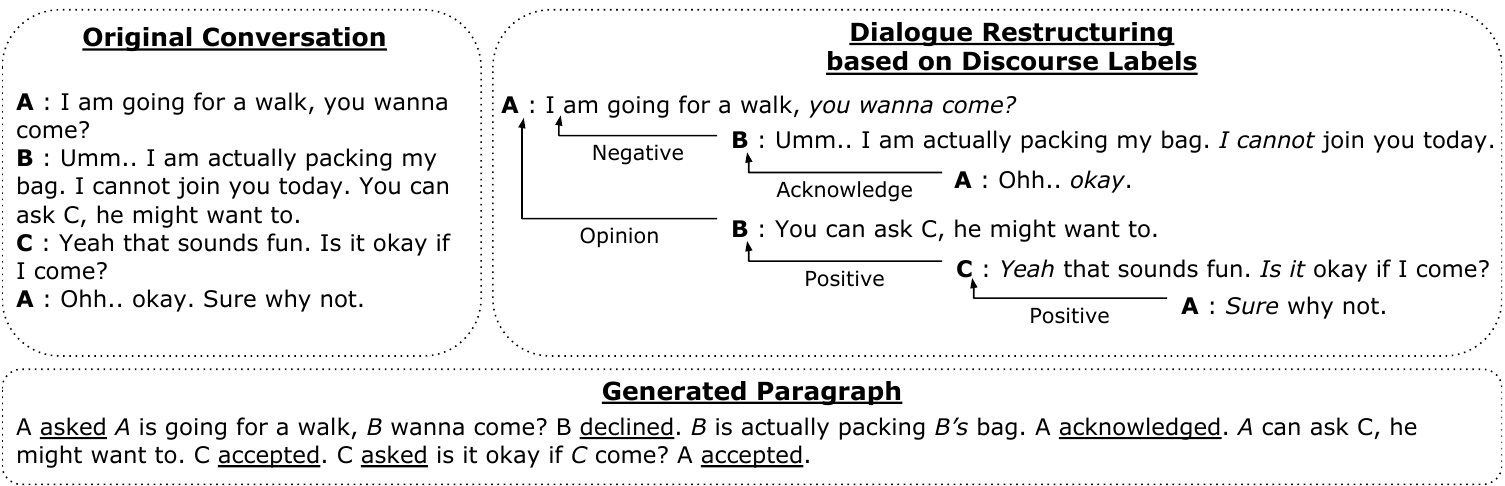}
\caption{Example conversation with restructuring using discourse labels and final generated paragraph. Underlined words in the generated paragraph denote dialog tags used and italic words denote anaphora resolution.}
\label{fig:discourse}
\end{figure*}

Real-life conversations, specially in case of multi-speaker setting, are highly unstructured and at times informal. We propose that discourse relations contain relevant information that can help us model the internal structure present in a dialogue. 
In the subsections below, we provide step-by-step details on leveraging discourse labels to convert a dialogue into a well-structured document.




\subsection{Discourse Labelling}
Discourse labelling is providing labels to every utterance present in the dialog by exploiting the inherent discourse structure of conversations.
Since conversations can be treated as sequence of utterances, providing discourse labels for each utterance in the conversation can be considered a sequence labelling task.
We use Conditional Random Fields (CRF) \cite{CRFsuite} in this paper, which has been recognized as a good baseline for discourse labelling \cite{kumar2017dialogue, ji2016latent, lee2016sequential}. We use both the word tokens and their POS tags (created using NLTK library \cite{loper2002nltk}) as features for CRF. 

\subsection{Dialogue Restructuring}
Once we have discourse labels for all the utterances, we use them to restructure the conversation into a documents using the steps below. An example of the complete pipeline is shown in Figure \ref{fig:discourse}.

\textbf{Anaphora resolution:}
Conversations usually occur in first and second person while documents are written in third person. We use a standard anaphora resolution tool \cite{loper2002nltk} to replace first and second person pronouns.

\textbf{Cleaning Fillers:}
Dialogues are informal in nature and can contain utterances that do not contribute to its meaning.
In fact, a major portion of dialogues, especially spoken conversations, contain fillers that can be removed without losing any information. 
Most of the previous work uses hard coded list of words or phrases in order to clean such fillers. However, we instead use predicted discourse tags to clearly identify such utterances and remove them from the dialog.
This can help us remove not only filler words, but entire filler utterances that do not contribute to the conversation.
We provide a few discourse tags and their corresponding examples that we clean up during this step in Table \ref{tab:cleanup}.

\textbf{Dialogue Reorganization:}
Certain utterances in a dialogue might only be understood when brought together but lose their meaning when separated, for example, Yes-No or Wh questions together with their answers.
In the conversion from dialog to document, we use predicted discourse labels to identify such pairs and reorder them appropriately. For example, specific discourse labels like \textit{Opinion} and \textit{Yes-No Question} have corresponding labels like  \textit{Response Acknowledgement} and \textit{Reject} later in the dialog sequence, probably from the different speaker.
We pair them using a rule-based system which relies on discourse labels, speaker roles and similarity between utterances.

\textbf{Adding Dialogue Tags:} Meaning of certain dialogue utterances can be summed up with simpler dialogue tags. These utterances do not hold any direct meaning but indicates an action performed by the speaker.
For example, saying, "Yes", "Yes, offcourse" or "Yes, I can see that" can all be summed up as the speaker 'agreeing' with the precedent in question. This can help significantly simplify the flow of information. We use tags like 'accepted', 'declined', 'acknowledged', etc. to represent corresponding discourse labels like \textit{Yes-Response}, \textit{No-Response}, \textit{Response Acknowledgement}, etc.


\begin{table}
\centering
\scriptsize
\setlength\tabcolsep{4pt}
\begin{tabular}{|c|c|}
\hline 
\textbf{Discourse labels} & \textbf{Example dialogues} \\
\hline
Conventional-opening & \textit{"How are you?"} \\
Conventional-closing & \textit{"Bye"} \\
Uninterpretable & \textit{"But, uh, yeah"} \\
Abandoned & \textit{"So, --"} \\
Hedge & \textit{"I'm not an expert but"} \\
Repeat-phrase & \textit{"Oh, fajitas"} \\
Non-understanding & \textit{"Excuse me?"} \\
\hline
\end{tabular}
\caption{Discourse labels for fillers with examples.}
\label{tab:cleanup}
\end{table}

\subsection{Document Summarization} 
Using our liberty to choose any document summarization tool, we experiment with 2 commonly used and publicly available models, pointer-generator networks \cite{see2017get}\footnote{github.com/abisee/pointer-generator} and transformer-based model BART \cite{lewis2019bart}\footnote{github.com/pytorch/fairseq/tree/master/examples/bart}. 






\section{Experiments}
\label{sec:experiments}

\begin{table*}
\centering
\scriptsize
\setlength\tabcolsep{2.8pt}
\begin{tabular}{|c||ccc|ccc|ccc||ccc|ccc|ccc|}
\hline 
\textbf{Method} & \multicolumn{9}{|c||}{\textbf{ICSI}} & \multicolumn{9}{|c|}{\textbf{AMI}} \\
\cline{2-19}
 & \multicolumn{3}{|c|}{\textbf{ROGUE-1}} & \multicolumn{3}{|c|}{\textbf{ROGUE-2}} & \multicolumn{3}{|c||}{\textbf{ROGUE-SU}} & \multicolumn{3}{|c|}{\textbf{ROGUE-1}} & \multicolumn{3}{|c|}{\textbf{ROGUE-2}} & \multicolumn{3}{|c|}{\textbf{ROGUE-SU}} \\
\hline
 & \textbf{R} & \textbf{P} & \textbf{F1} & \textbf{R} & \textbf{P} & \textbf{F1} & \textbf{R} & \textbf{P} & \textbf{F1} & \textbf{R} & \textbf{P} & \textbf{F1} & \textbf{R} & \textbf{P} & \textbf{F1} & \textbf{R} & \textbf{P} & \textbf{F1} \\
 \hline
Longest Greedy & 35.57 & 26.74 & 30.23 & 4.84 & 3.88 & 4.27 & 13.09 & 9.46 & 10.90 & 37.31 & 30.93 & 33.35 & 5.77 & 4.71 & 5.11 & 13.79 & 11.11 & 12.15 \\
\hline
TextRank & 34.89 & 26.33 & 29.70 & 4.60 & 3.74 & 4.09 & 12.42 & 9.43 & 10.64 & 39.55 & 32.60 & 35.25 & 7.67 & 6.43 & 6.90 & 14.87 & 12.87 & 13.62 \\
\hline
CoreRank Submodular & 35.22 & 26.34 & \textbf{29.82} & 4.36 & 3.76 & \textbf{4.00} & 12.11 & 9.58 & \textbf{10.61} & 41.14 & 32.93 & \textbf{36.13} & 8.06 & 6.88 & \textbf{7.33} & 14.84 & 13.91 & \textbf{14.18} \\
\Xhline{3\arrayrulewidth}
\cite{shang2018unsupervised} & 35.95 & 27.00 & \textbf{30.52} & 4.64 & 3.64 & \textbf{4.04} & 12.43 & 9.23 & \textbf{10.50} & 42.43 & 35.01 & \textbf{37.86} & 8.72 & 7.29 & \textbf{7.84} & 16.19 & 13.76 & \textbf{14.71} \\
\Xhline{3\arrayrulewidth}
PGN \cite{see2017get} & 37.20 & 21.77 & 27.26 & 4.47 & 3.08 & 3.62 & 11.60 & 7.80 & \textbf{9.26} & 38.20 & 27.08 & 31.27 & 6.32 & 4.57 & 5.23 & 13.76 & 9.82 & 11.30 \\
\hline
PGN + Discourse & 34.35 & 22.95 & \textbf{27.35} & 4.20 & 3.25 & \textbf{3.64} & 10.63 & 8.12 & 9.15 & 39.27 & 31.25 & \textbf{34.24} & 7.42 & 5.95 & \textbf{6.50} & 15.00 & 11.86 & \textbf{13.03} \\
\Xhline{3\arrayrulewidth}
BART \cite{lewis2019bart} & 33.31 & 27.96 & 30.11 & 4.39 & 4.18 & 4.23 & 10.38 & 9.76 & 9.94 & 34.64 & 30.25 & 31.84 & 6.80 & 5.83 & 6.19 & 12.95 & 10.77 & 11.58 \\
\hline
BART + Discourse & 34.64 & 30.25 & \textbf{31.84} & 6.80 & 5.83 & \textbf{6.19} & 12.95 & 10.77 & \textbf{11.58} & 38.67 & 33.72 & \textbf{35.41} & 8.04 & 6.79 & \textbf{7.24} & 15.14 & 12.20 & \textbf{13.27} \\
\hline
\end{tabular}
\caption{Results of various methods on the ICSI and AMI test set, macro-averaged for 350 and 450 word summaries}
\label{tab:ami}
\end{table*}

\subsection{Datasets and Metric}

We use 2 different datasets to explore the adaptability of our model across domains, i.e., the AMI \cite{carletta2005ami} and ICSI \cite{janin2003icsi} meeting corpus.
The Switchboard Dialog Act Corpus \cite{jurafsky1997switchboard}, which contains 1115 conversations with 205,000 utterances and 43 different discourse tags, was used to train the CRF. The out-of-the-box summarization methods were originally trained on CNN/DailyMail dataset \cite{hermann2015teaching, nallapati2016abstractive}, which contains over 300K online news articles (781 tokens on average) paired with multi-sentence abstractive summaries (3.75 sentences or 56 tokens on average). 

Since the document summarization models were trained on smaller documents, we chunk the input dialogue into smaller pieces and summarize each chunk separately before combining the outputs. We create chunks of length 800 tokens, keeping in mind the original CNN/Dailymail dataset length of 781 tokens on average.

We replicate the evaluation settings proposed by \cite{shang2018unsupervised} as is, allowing us to borrow 3 simple extractive summarization baselines from them and also directly compare our performance with them. Please refer to the supplementary file, or the work by \citet{shang2018unsupervised}, for more details regarding the evaluation setup.

\subsection{Results and Interpretation}

Results of our framework on the AMI and ICSI corpus, along with other baselines, are present in Table \ref{tab:ami}. When compared with base document summarization, our method significantly improves the performance when discourse based dialogue restructuring is applied. This can be explained as the document summarizer was originally trained on structured input news articles, and thus our transformation of the input dialogue helps bridge the gap between the two.
Also, it can be noted that for a better final performance, we require a powerful base document summarizer. Since BART is a better document summarizer then PGN, the final model performance also follows the same trend.

Although it is unfair to compare ROGUE scores directly between our method and the baselines, as the baselines are either zero-shot extractive summarization methods or an unsupervised abstractive summarization method \cite{shang2018unsupervised}, while we are doing zero-shot abstractive summarization. But for reference, we show that we can perform competitively against these methods. Specially for ICSI, we even achieve better scores than these baselines when using a good document summarizer.
\citet{zhao2019abstractive} trained a neural model on the AMI test set, and was able to improve the performance by 7 ROGUE points over PGN (the evaluation setup was different), while we are able to improve up to 3 ROGUE points without the requirement of any training data.

Finally, since ROGUE scores are known to favor extractive summaries, these scores alone cannot be a true judge of the model's capabilities. We also provide various real-life conversations in supplementary file, picked randomly from the internet, along with generated summaries, in order to provide a qualitative comparison of our method.
\section{Conclusion and Future Work}
\label{sec:conclusion}

We proposed a zero-shot abstractive dialogue summarization method which uses existing state-of-the-art discourse labeling and document summarization models without any additional training. While existing work in this field focuses on training specific models for every domain, our model can be generalised to multiple domains and our novelty lies in simplicity of the pipeline which allows the user to adapt easily available document summarizers for dialogues.
Our method can help create more intuitive summaries without any additional training cost whatsoever, which is backed with both quantitative as well as empirical evidence.
Given the unavailability of domain specific dialog data, our work shows the importance of converting dialogues, which contain significant noise and does not follow a linear structure, into documents.

While we propose some primitive methods to perform this conversion, which can be improved by replacing them with more robust predictive methods, we do bring forward the merit of research in this direction and the need to move away from the existing trend of training domain-specific models for improving their practical application. Also, as evident by our results, the final performance of our model is dependent on the out-of-the-box models used, and thus improvement in those directions of research will also directly fuel further improvement in our model's performance.



\bibliography{acl2020}
\bibliographystyle{acl_natbib}

\section{Background}

We provide more details on the document summarization models and datasets used, as well the evaluation setup for our experiments. We also provide 3 real life conversations taken from diverse sources, in order to emphasize the generalization capabilities of our model.

\subsection{Document Summarization models}

Pointer-generator networks by \citet{see2017get} is the latest in a long line of RNN-based seq2seq models for document summarization. While abstractive summarization models usually generate summaries from scratch, this particular model concurrently uses a copy mechanism, allowing the model to copy words directly from the source text. This aids in accurate reproduction of information while retaining the ability to produce novel words through the generator. The authors also introduce coverage mechanism to keep track of what has been summarized, which discourages repetition.

Transformer-based models have gained a lot of momentum in the last few years and have been shown to be significantly more powerful than their RNN-based counterparts. An important advantage of such models is the unsupervised pre-training on large corpus of data from the internet. Transformers are inherently known to be robust to small noises present in the input, but BART \cite{lewis2019bart} takes it even a step further by explicitly introducing noisy data during training and thus improving the model's robustness.

\subsection{Datasets}

The AMI meeting corpus \cite{carletta2005ami} contains targeted meetings between multiple speakers, with a pre-defined agenda. It contains 20 conversation in the test set, each with one reference abstractive summary of length 290 words on average. The ICSI meeting corpus \cite{janin2003icsi} is also a multi-speaker dataset with 6 conversations in its test set. Each conversation in ICSI contains three reference summaries of length 220, 220 and 670 words on average.

\subsection{Evaluation setup}
For a fair comparison with other baseline methods, we follow the same evaluation protocol used by \cite{shang2018unsupervised}. We use ROGUE-1, 2 \& SU scores \cite{lin-2004-rouge} for evaluation, a commonly used metric used to judge the quality of summaries.
We report the macro average ROGUE scores of 2 summaries of length 350 and 450 words, which are matched against the reference summary in AMI and all three reference summaries in ICSI.
We keep the beam search width = 8 for all these experiments.
We also borrow 3 simple zero-shot extractive approaches from their work for reference. These are \textbf{Longest Greedy} \cite{riedhammer2008packing}, \textbf{TextRank} \cite{mihalcea2004textrank} and \textbf{CoreRank Submodular} \cite{tixier2017combining}.

\section{Real life Conversation Examples}

We use 3 real life conversation examples to understand the extend of generalization of our model (see Tables \ref{tab:real1}, \ref{tab:real2} and \ref{tab:real3} in the following pages). BART only summaries refer to directly using the out-of-the-box document summarizer on the conversations, while BART + Discourse refers to the summaries created by our method. All the tables are present from page 3 onwards.

The first conversation is from an NPR podcast, whose transcript is available online \footnote{https://www.npr.org/2020/05/23/861577391/long-term-symptoms-of-covid-19}. It is an example of interview type conversation between two speakers, where one speaker is asking questions and the other speaker is responding. The out-of-the-box document summarizer is not used to such an exchange and thus only picks information from statements (and not questions) by various speakers. However, after using discourse relations to restructure the document, we are able to convert even the questions asked by the interviewer into proper format and BART + Discourse summarizer is thus able to extract important information from the complete conversation, instead of just the statements.

Next, we picked a spoken conversation from Better English, an online platform for user to improve their grasp on the English language \footnote{https://www.betteratenglish.com/real-english-conversations-lori-scores-a-years-supply-of-toilet-paper}. This conversation is more informal than the previous one. It is actually more of a monologue than a conversation, with the other speaker occasionally chiming in. The language is extremely informal. It can be noted that using a document summarizer directly on such a conversation created a summary that isn't readable and contains incomplete sentences. However, after using discourse relation to clean filler utterances, the final summary created by our method looks more structured and proper.

Finally, we present a conversation between few of our colleagues (names changed) who are non-native English speakers. The conversation contains a lot of technical terms from a specific domain specialisation that the document summmarizer has probably not seen before. This conversation also contains more than two speakers and the length of a single utterance is also smaller than the last two conversations. The generated summary directly from the document summarizer contains both a question and its answer, while only one of those two should have been enough to give the complete information. This does not happen in our summary, as we are able to help the summarizer connect the question with its answer. This gives our method space to add more information, like the project on which Liam is working, which went completely unnoticed by the document summarizer earlier.

\begin{table*}
\centering
\begin{tabular}{|p{0.95\linewidth}|}
\hline 
\textbf{Real Life Conversation Example \#1} \\ 
\hline
\textbf{Mara}:That day, I was feeling, you know, myself, felt really good. And then the next day, on April 17, I woke up, and I felt hot and feverish. Then unfortunately, the next morning, on April 18, I woke up, and it felt like, you know, an elephant was sitting on my chest. And it was pretty scary. \\
\textbf{Martin}:I want to quote a part of your piece where you make it clear why you wrote it. You said you want Americans to understand that this virus is making otherwise young, healthy people very, very sick. You want them to know this is no flu. Do you get the sense that there are people who still don't understand the seriousness of this? \\
\textbf{Mara}:Yes. Yes. I think we know now that this virus can be extremely aggressive with even younger healthy people. I just wanted Americans to understand that they're rolling the dice. When you get something like this, you don't know how your body is going to respond. \\
\textbf{Martin}:I wanted to ask - you know, getting back to the you part of this, you mentioned that you're still recovering, you know, all these weeks later. You've - have pneumonia and restrictive airways and... \\
\textbf{Mara}:Yeah. \\
\textbf{Martin}:...Reactive airways - asthma, as it were. Apart from the physical symptom, do you think this has - experience has changed you in some way? I wonder if you feel like this will stick with you after you recover physically. \\
\textbf{Mara}:Well, it definitely will. I don't know all the ways yet. I'm still kind of going through it. But, you know, I thought a lot when I was really sick about the Americans who got sick the same time that I did and, you know, didn't recover. I think I'm going to be thinking a lot about how to do right by them by living my life to the fullest and trying to pay it forward. And I got a lot of help when I was very sick. I'm still getting a lot of help, so I just want to make sure that, you know, I can be a part of helping others who may not have the same privileges or advantages or family and friends and support that I did. \\
\textbf{Martin}:Mara Gay with All Things Considered host Michel Martin . \\
\hline
\textbf{Generated Summary - BART only} \\
\hline
Mara Gay with All Things Considered host Michel Martin. Mara said I just wanted Americans to understand that they are rolling the dice. When you get something like this, you do not know how your body is going to respond. Mara said I thought a lot when I was really sick about the Americans who got sick. \\
\hline
\textbf{Generated Summary - BART + Discourse} \\
\hline
That day, Mara was feeling, myself, felt really good. And then the next day, on April 17, Mara woke up, and Mara felt hot and feverish. Mara want Americans to understand that this virus is making otherwise young, healthy people very, very sick. \\
\hline
\end{tabular}
\caption{Real Life Conversation Example \#1}
\label{tab:real1}
\end{table*}

\begin{table*}
\centering
\begin{tabular}{|p{0.95\linewidth}|}
\hline 
\textbf{Real Life Conversation Example \#2} \\ 
\hline
\textbf{Lori}:Yeah, something kind of funny happened to me when I was shopping for office supplies today. \\
\textbf{Andy}:OK, what happened? \\
\textbf{Lori}:Well, my boss had, had given me a list of office supplies to buy on my way home from a teaching gig, because I drive right past the office supply shop. And I’m always happy to do it, ’cause, as you know, I LOVE office supplies — it’s almost like my, my “office-supply porn” — and…I had a whole list of things to buy. And when I got up to the register and the clerk was ringing me up, the total came to over a thousand Swedish crowns. Which is not a problem, I mean, they just just send us an invoice; it wasn’t like I had to worry about money. But then he said, “Because you spent so much money here today, you can go pick one of those rolls of toilet paper over there.” \\
\textbf{Andy}:Toilet paper! \\
\textbf{Lori}:Yeah, toilet paper! And, I mean, we’re always happy to get free toilet paper; you know, it’s one of those useful things that, that, you know, a business has to buy… \\
\textbf{Andy}:You can never have too much. \\
\textbf{Lori}:Yeah, exactly. But the thing is, I looked at where he was pointing, and it was these HUGE, GIGANTIC, industrial-sized packages, all shrink-wrapped in plastic, of toilet paper…I mean, it was HUGE, I could NOT BELIEVE that I was getting one for free. \\
\textbf{Andy}:OK, like a year’s supply of toilet paper. \\
\textbf{Lori}:At least. I’m serious! When…standing on end, the thing comes up almost to my chest. I mean, it’s huge. I, I forgot to count the rolls, but it was…it had to be…maybe… At least 20 packs of six rolls each. \\
\textbf{Andy}:Wow. \\
\textbf{Lori}:Seriously, it was one big, honking supply of toilet paper. \\
\textbf{Andy}:And this is free? \\
\textbf{Lori}:Yeah, free just because I’d spent, you know, in one, you know, one purchase, we had spent over a thousand crowns. And I, but I could not believe they were giving away for free, and so I had to ask the guy, “Really? Are you kidding? You mean I get to take one of these?” And he was like, “Yeah, yeah.” I’m like, “No!” He was like, “Yeah!” “No!” And he says that, “You know, you can look, see the sign up above…it says…I can show you.” I’m like, “No no, it’s not that I don’t believe you, I just can’t believe you are giving away such a huge supply of toilet paper!” I was REALLY happy. And of course there was a line of Swedes standing waiting to pay for their things, and they were raising their eyebrows at me, you know, someone getting SO excited about getting a huge supply of toilet paper. But you know, I thought that was just a really really cool thing.. for or the company to do. ‘Cause say they had just given away one little pack, or two little packs. Like, oh, wow, 12 rolls of toilet paper. For spending a thousand crowns… \\
\textbf{Andy}:Yeah. No, free stuff is good! \\
\textbf{Lori}:Yeah, and the good news is, you know here at home we’re on our last roll… And because I scored this huge supply of toilet paper for my boss, she’s like, “Take some, take some!” And… \\
\textbf{Andy}:NOW I see why you’re so happy. \\
\textbf{Lori}:That’s why today I came home with that, you know, with… \\
\textbf{Andy}:Your arms full of toilet paper. \\
\textbf{Lori}:Exactly, exactly. \\
\hline
\textbf{Generated Summary - BART only} \\
\hline
Lori said something kind of funny happened to me when I was shopping for office supplies today. Lori said I was shopping for office supplies when my boss gave me a list of things to buy. I came home with that , I know , with…  your arms full of toilet paper.
 \\
\hline
\textbf{Generated Summary - BART + Discourse} \\
\hline
Lori's boss had given Lori a list of office supplies to buy on Lori's way home from a teaching gig. The total came to over a thousand Swedish crowns, which is not a problem. But then he said "Because you spent so much money here today, you can go pick one of those rolls of toilet paper. \\
\hline
\end{tabular}
\caption{Real Life Conversation Example \#2}
\label{tab:real2}
\end{table*}

\begin{table*}
\centering
\begin{tabular}{|p{0.95\linewidth}|}
\hline 
\textbf{Real Life Conversation Example \#3} \\ 
\hline
\textbf{John}:I think it is time to decide on a topic for the NLP project. \\
\textbf{Kevin}:You are right. Ummm... I am too confused. How can we read a paper, code it and improve the baseline within a month? This seems like too much to ask for. \\
\textbf{John}:Haha.. Don't worry. We can look online for a paper which comes with existing codebase and then directly jump on improving the baseline. I think finding a paper with code as well as dataset available will be a challenge. Why don't we try something on sentiment analysis?\\
\textbf{Kevin}:Yeah! That seems cool. But I believe working on plain sentiment analysis will be boring since we already have done an assignment on it. Are you familiar with generative language models? \\
\textbf{John}:Yes. They are really interesting. But Ummm... This is gonna be tough. I think this will require a lot of hardwork. Given that this semester is fully loaded, why do you want to try a new thing? \\
\textbf{Kevin}:I too believe generative models are tough. They will be difficult to train and implement. But isn't trying new things is what we are here for. I am up for this. \\
\textbf{Liam}:Hi guys. What are you doing? \\
\textbf{Kevin}:Hi Liam. \\
\textbf{John}:Hey Liam. We were just discussing about our NLP course project this semester and I think we have decided the domain. It is going to be on Generative Model for sure now. \\
\textbf{Kevin}:Liam, What are you planning of doing? \\
\textbf{Liam}:I am planning of doing something to remove bias words from Wikipedia articles. \\
\textbf{Kevin}:Wow! That sounds great. We discussed so much in this chat. John, why don't we have something that can summarize this discussion? \\
\textbf{John}:You mean like an auto dialogue summarization based on generative language modelling? \\
\textbf{Kevin}:Yes, exactly! So I believe it's time to get started. \\
\hline
\textbf{Generated Summary - BART only} \\
\hline
John said  I think it is time to decide on a topic for the NLP project. Kevin said  How can we read a paper, code it and improve the baseline within a month ? John said  We can look online for a paper which comes with existing codebase and then directly jump on improving the baseline. \\
\hline
\textbf{Generated Summary - BART + Discourse} \\
\hline
John think it is time to decide on a topic for the NLP project. Kevin believe working on plain sentiment analysis will be boring. John think finding a paper with code as well as dataset available will be a challenge. Liam is planning of doing something to remove bias words from Wikipedia articles. \\
\hline
\end{tabular}
\caption{Real Life Conversation Example \#3}
\label{tab:real3}
\end{table*}

\end{document}